\documentclass[11pt,a4paper,twocolumn]{article}
\usepackage{mathptmx}

\usepackage[a4paper, margin=1in]{geometry}
\usepackage{graphicx}
\usepackage{subcaption}     
\usepackage{amsmath}
\usepackage{amsfonts}
\usepackage{amssymb}
\usepackage{url}
\usepackage{hyperref}
\hypersetup{
    colorlinks=true,
    linkcolor=blue,
    filecolor=magenta,
    urlcolor=cyan,
}

\begin{document}

\newcommand{\etal}{\textit{et al. }}
\providecommand{\keywords}[1]{\textbf{\textit{Keywords - }} #1}

\title{Image classification in frequency domain with 2SReLU: a second harmonics superposition activation function}

\author{
  Thomio Watanabe\\
  \texttt{University of Sao Paulo}\\
  \textit{thomio.watanabe@usp.br}
  \and
  Denis F. Wolf\\
  \texttt{University of Sao Paulo}\\
  \textit{denis@icmc.usp.br}
}
\date{}

\maketitle

\section*{Abstract}

Deep Convolutional Neural Networks are able to identify complex patterns and perform tasks with super-human capabilities.
However, besides the exceptional results, they are not completely understood and it is still impractical to hand-engineer similar solutions.
In this work, an image classification Convolutional Neural Network and its building blocks are described from a frequency domain perspective.
Some network layers have established counterparts in the frequency domain like the convolutional and pooling layers.
We propose the 2SReLU layer, a novel non-linear activation function that preserves high frequency components in deep networks.
It is demonstrated that in the frequency domain it is possible to achieve competitive results without using the computationally costly convolution operation.
A source code implementation in PyTorch is provided at: \url{https://gitlab.com/thomio/2srelu}.
\keywords{
  Neural networks,
  image classification,
  frequency domain,
  activation function.
}

\section{Introduction}


After the deep learning breakthrough, AlexNet \cite{krizhevsky2012imagenet}, most solutions for computer vision problems are based in deep Convolutional Neural Networks (CNN).
The previous handcrafted feature descriptors were replaced by a single model that automatically learns how to generate the best features to solve a problem.
Although these models are being widely employed, they are still considered black boxes where it is still not clear the purpose of each feature extractor \cite{ribeiro2016should, lundberg2017unified, rudin2019stop}.

Deep CNNs directly utilize image data which are described by spatial domain variables.
In other words, the information is represented in the spatial domain.
But this information can also be represented in the frequency domain, where the data is described by a composition of frequency components.
The frequency domain representation presents the same information in a different manner and also simplifies the application of specific operations, like filtering.

Each CNN layer performs an operation that can be separately described in the frequency domain.
The convolution is the main layer in CNNs, and it is defined in the frequency domain by the convolution theorem.
The convolution theorem states the convolution is equivalent to a pointwise multiplication in the frequency domain.
Some other layers, like the pooling layer, have been studied in the frequency domain but the ReLU layer is still an open problem.
Finding all layers equivalents in the frequency domain will enable researches to train networks with frequency domain data and it will provide extra information to understand how deep neural networks function.

One benefit of a frequency domain neural network is the reduction of the computational complexity.
This happens because the 2D convolution operation presents a high computational complexity, $O(N^2 k^2)$, while the pointwise product operation is lower, $O(N^2)$.
Although the pointwise product decreases the computational complexity it increases the number of layer parameters (memory) required to perform the operation.
A single convolution presents a number of parameters equal to the kernel size, $k^2$.
While the pointwise multiplication requires a number of parameters (weights) equal to the input signal size, $N^2$.

This work addresses the development of convolutional neural networks for image classification in the frequency domain.
The related work section presents an overview of the state-of-the-art solutions for frequency domain layers.
Each layer operation is analyzed in the frequency domain illustrating their interaction inside a network.
It is pointed out the spectral pooling high frequency components removal problem while a novel non-linear activation function that directly addresses the spectral pooling deficiency is proposed.
A set of experiments have been performed to assess the proposed layer and also to compare it with a similar solution.


\section{Related Work}

\subsection{Convolution}

The convolution is a mathematical operation defined by Equation \ref{eq_convolution}.
In signal processing the convolution is employed to filter input signals separating its frequency components.
The convolution is being widely employed in neural networks as the main building block of convolutional neural networks.

\begin{equation}
f(x) \ast g(x) = \int_{-\infty}^{\infty} f(\tau)g(x-\tau)d\tau
\label{eq_convolution}
\end{equation}

In frequency domain analysis, the convolution theorem states that:

\newtheorem{theorem}{Theorem}
\begin{theorem}
the Fourier transform of a convolution of two signals is the Hadamard pointwise product of their Fourier transforms:
\begin{eqnarray}
\label{eq_convolution_theorem}
f(x) \ast g(x)  \xrightarrow{\mathfrak{F}} F(\omega) \cdot G(\omega)
\end{eqnarray}
\end{theorem}

This theorem can be applied to time-domain signals and also for space domain signals like images.


Specifying high-band, low-band and pass-band filters in the frequency domain is a straightforward process.
These filters can be designed in the frequency domain and then transformed to its original domain with the inverse Fourier transform.

Recent work has investigated the convolution theorem application in deep neural networks, \cite{rippel2015spectral, wang2016combining, pratt2017fcnn, ayat2019spectral}.
In common, they argue that replacing the convolutional layer by a pointwise product substantially decreases the network computational complexity speeding up the network training and inference.

\subsection{Normalization}

Normalization layers are widely employed in deep neural network.
These layers provide numerical stability, regularizing the network while reducing the learning time.
Two common normalization layers are the Batch Normalization (BN) \cite{ioffe2015bn} which is the main reference for normalization layers, and the Group Normalization (GN) \cite{wu2018group}, which is more stable for small batch sizes.

To normalize complex numbers, Trabelsi \etal \cite{trabelsi2018deep} separate the complex real and imaginary parts and divide the feature maps with the square root of the $2\times 2$ covariance matrix.
The authors state the normalization should ensure equal variance in the real and imaginary components.

Ayat \etal \cite{ayat2019spectral} separate the batch normalization layer in two parts: the scaling and shifting operations.
The shifting operation is the mean value removal and it is equivalent to the DC component removal.
The scaling operation is the division by the standard deviation which was considered equivalent to the division by the frequency domain standard deviation.

\subsection{ReLU}

The Rectified Linear Units (ReLU) layer \cite{nair2010rectified} is a simple non-linear activation function defined by Equation \ref{eq_relu}.
Negative feature maps values are zeroed in the forward pass and during the backpropagation their gradients are blocked.

\begin{equation}
  ReLU(x) = max(x,0)
  \label{eq_relu}
\end{equation}

Despite being a simple function its purpose inside neural networks is still not clear (add non-linearities).
It has a strong biological motivation but a shallow mathematical justification.
Some non-linear activation functions for complex numbers are presented in this section, following the naming convention defined by Trabelsi \etal \cite{trabelsi2018deep}.

\subsubsection{modReLU}

The modReLU activation function \cite{arjovsky2016unitary} alters the magnitude value of input signals by a learnable parameter $b$.
If the input magnitude and $b$ summation is negative the output is zeroed.

\begin{align}
  modReLU(x) = ReLU(|x| +b) e^{i\theta} =\\ \nonumber
  \begin{cases}
     (|x|+b)e^{i\theta} &\text{if} \quad |x| + b \geq 0\\
      0 &\text{otherwise}
  \end{cases}
  \label{eq_modrelu}
\end{align}

\subsubsection{zReLU}
The zReLU activation function \cite{guberman2016complex} only keeps the complex values where the phase belongs to the first quadrant $\Re{(x)} > 0$ and $\Im{(x)} > 0$.

\begin{equation}
  zReLU(x) =
  \begin{cases}
    x \quad &\text{if} \quad \theta \in [0, \pi /2]\\
    0 &\text{otherwise}
  \end{cases}
\end{equation}

\subsubsection{CReLU}

The CReLU was presented by Trabelsi \etal \cite{trabelsi2018deep} and directly applies the ReLU function in the complex input real and imaginary components.

\begin{equation}
  \mathbb{C}ReLU(x) = ReLU( \Re(x) ) + i ReLU( \Im(x) )
\end{equation}

All these previous activation functions (modReLU, CReLU and zReLU) apply non-linearities in each individual frequency component.
In other words, different frequency components values do not interfere each other.
In Section \ref{sec_freq_analysis}, is explained why this approach does not represent the actual ReLU operation.

\subsubsection{Spectral ReLU}

Recently, Ayat \etal \cite{ayat2019spectral} proposed a spectral ReLU (SReLU) layer.
The authors model the ReLU layer in the frequency domain with the convolution operation.
Their idea is to approximate the ReLU by a displaced parabola.
Basically, the SReLU convolve the input feature map with itself, according to Equation \ref{eq_srelu}.

\begin{equation}
  srelu(X_i) = c_2 (X_i \ast X_i) + c_1 X_i + c_0
  \label{eq_srelu}
\end{equation}

\begin{figure}
   \centering
   \includegraphics[width=\columnwidth]{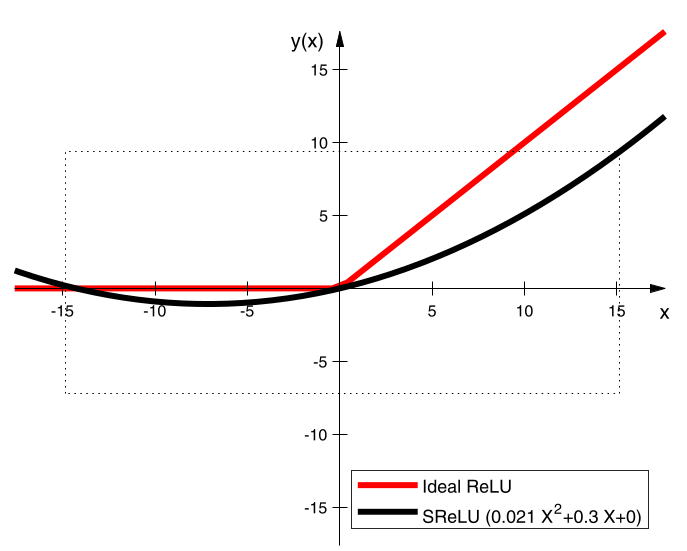}
   \caption{ ReLU and SReLU illustration in spatial domain. A convolution in frequency domain is approximated by a parabola in the space domain. Image from Ayat \etal \cite{ayat2019spectral}. }
   \label{fig_srelu}
\end{figure}

The convolution in the frequency domain is equivalent to a pixel-wise multiplication in the spatial domain, and the SReLU can be approximated by a parabola similar to Figure \ref{fig_srelu}.
The coefficients $c_0$, $c_1$ and $c_2$ are manually defined for each problem and the input features must be limited between a range, $(-15,15)$ in Figure \ref{fig_srelu}.

One of the main arguments to create a neural network in the frequency domain is to completely remove the convolution operation from the model, reducing the computational complexity.
Replacing the ReLU operation with a convolution goes in the opposite direction and it may increase the computational cost of the model.



\subsection{Pooling}

The pooling operation reduces the feature maps resolution increasing the network receptive field.
In the frequency domain, this operation is equivalent to a low-pass filter that is explained in most image processing textbooks.

\subsubsection{Fourier Spectral Pooling}

The Fourier spectral pooling proposed by Rippel \etal \cite{rippel2015spectral} replaced the spatial pooling operation by truncating the features representation in the frequency domain.
This approach preserves more information than spatial pooling strategies and allows a flexible output size adjustment.

\begin{figure}
   \centering
   \includegraphics[width=1.\columnwidth]{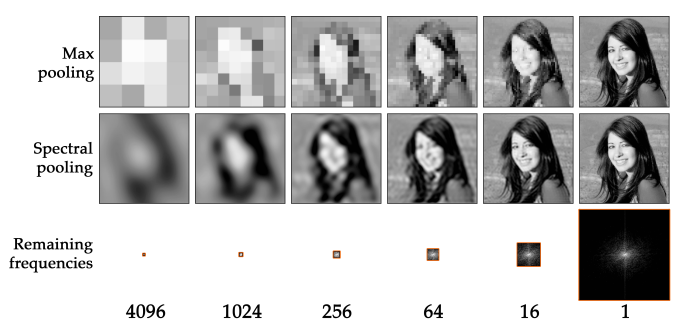}
   \caption{ Image resolution reduction with max pooling and spectral pooling. Spectral pooling preserves more information than max pooling. Image from Rippel \etal \cite{rippel2015spectral}. }
   \label{fig_spool}
\end{figure}

Pooling layers perform an operation equivalent to a low-pass filtering.
Performing the operation is the frequency domain preserves more information than in the spatial domain.
This happens because the power spectra of natural images is concentrated on lower frequencies and the spectral pooling can precisely separate low frequencies from high frequencies.
This observation is illustrated in Figure \ref{fig_spool}.

High frequency components are directly related to details and noise.
In images, high frequency components are important to define contours and borders.
However, in some tasks like image classification, object contours are not a critical information.

Rippel \etal \cite{rippel2015spectral} were not able to create a network in the frequency domain because there was no ReLU layer (operation) equivalent in the frequency domain.
To perform experiments the authors actively computed the Discrete Fourier Transform (DFT) and its inverse, which is a viable solution but highly increases the computation time.
The authors conclude the spectral pooling helps the network to converge with a smaller number of iterations.

Since the pooling operation is equivalent to a low-pass filtering, some argue it should be replaced by convolutions with non-unitary stride.
In this case the network should be able to learn the filter parameters.
Some work, like ResNet \cite{he2016deep}, present networks with both strategies.

\subsubsection{DCT-based Pooling}

The Discrete Cosine Transform (DCT) is also a frequency domain transform which is being utilized in pooling layers \cite{smith2018discrete, ryu2018dft, xu2019dct}.
These papers propose modifications to preserve more information or reduce the layer computational cost.
Experimental results demonstrate that DFT pooling layers improves the classification performance when compared with traditional pooling layers.

These methods are conceptually similar, they transform the feature maps in the frequency domain and they create a low-pass filter to separate low frequency components from high frequency components.
The DCT advantage over the DFT is that it has a lower computational cost and its output presents only real numbers.

\subsubsection{Hartley Spectral Pooling}
Based on Fourier spectral pooling \cite{rippel2015spectral}, Hao and Ma \cite{zhang2018hartley} proposed the Hartley spectral pooling.
This layer employs the Hartley transform which is also a frequency transform. 
The Hartley transform does not present imaginary numbers and prevents the use of complex arithmetic.
This layer has a lower computational cost than the Fourier spectral pooling and the authors demonstrate it decreases the convergence time and provides a higher classification accuracy than its counterpart max pooling.


\section{Frequency Domain}
\label{sec_freq_analysis}

The Fourier transform is employed to represent signals and systems in the frequency domain.
In this analysis, the signals are 2D images and the systems are deep neural network layers.
In this space-frequency duality each spatial operation has an equivalent in the frequency domain.

When representing signals in the frequency domain is possible to observe the problem from a different perspective.
Frequency transforms are a powerful tool to understand complex models and they also provide means to perform operations that might be impractical in the original domain.


\subsection{Factored Convolutions}

The factored convolutions were introduced by the VGG network \cite{simonyan2014very}.
It consists of stacking blocks of convolutional layers and activation functions before a pooling layer.
This arrangement generates more complex features while increasing the receptive field of the network.
Stacking convolutional layers is equivalent of using a single convolution layer with a larger kernel.
For instance, two (3x3) convolutional layers is equivalent to one (5x5) convolutional layer and three (3x3) convolutional layers is equivalent to one (7x7) convolutional layer.

In the frequency domain, the receptive field concept does not exist because the input data is not a natural image anymore.
In other words, the input data is not spatially correlated.
Also, the filter in the frequency domain encloses all frequencies components.

\subsection{Batch Normalization}


The batch normalization (BN) layer \cite{ioffe2015bn} is one of the most important normalization layers and it is widely employed in classification, detection and segmentation networks. 
This layer normalizes the input feature maps by its batch standard deviation $\sigma$ and removes its mean value, $\mu$, according to Equations \ref{eq_bn} and \ref{eq_bn_params}.



\begin{align}
  \label{eq_bn}
  \hat{x} &= \frac{x - \mu}{\sqrt{\sigma^2 + \varepsilon}}
\end{align}

\begin{align}
  \label{eq_bn_params}
  \mu = \frac{1}{m} \sum^{m}_{i=1} x_{i} \qquad
  \sigma^2 = \frac{1}{m} \sum^{m}_{i=1} \left( x_{i} - \mu \right)^2 
\end{align}

The BN also have two learnable parameters $\gamma$ and $\beta$, utilized to scale and to shift the input data, Equation \ref{eq_bn_learnable}.

\begin{equation}
  \label{eq_bn_learnable}
  y = \gamma \hat{x} + \beta
\end{equation}

In summary, the layer tries to learn what are the optimal scaling and shifting parameters.
These two learnable parameters are optional in most deep learning frameworks and they will be neglected in this evaluation.

The Equation \ref{eq_bn} carries a prior about data processing.
In image processing the mean value is associated with an image brightness and removing this information implies the solution is independent of this feature.
Normalizing the input data by its standard deviation produces a distribution with unit variance.
Therefore, if the input data can be modeled by a normal distribution, 68\% of the data is within the $[-1,1]$ interval.
The BN bounds the feature maps and consequently the network weights to small values, increasing numerical stability and decreasing the number of network training iterations.

In the frequency domain, removing the mean value is equivalent to zeroing the DC frequency component.
It is a simple operation and can be described by a high-pass filter.
Regarding the input division by the standard deviation, an equivalent operation in the frequency domain was not identified.
However, the same operation in both domains would provide numerical stability during the network training process.

\subsection{ReLU}

To understand what the ReLU operation does in the frequency domain, some examples with sine waves and their representation in the frequency domain were examined.

To prevent frequency aliasing and consequently information loss in the frequency domain, the sampling frequency must satisfy the Nyquist-Shannon sampling theorem, Equation \ref{eq_nyquist}.
The sampling frequency $F_s$ must be higher than twice the input signal highest frequency component $\mu_{max}$.

\begin{equation}
  F_{s} > 2\mu_{max}
  \label{eq_nyquist}
\end{equation}

In the following examples we set $F_s = 100$ in a unitary cinterval, resulting in one hundred samples, $N=100$.
For this setup the input higher frequency must be smaller than $\mu_{max} = 50$.

\begin{figure}
  \centering
  \includegraphics[width=\columnwidth]{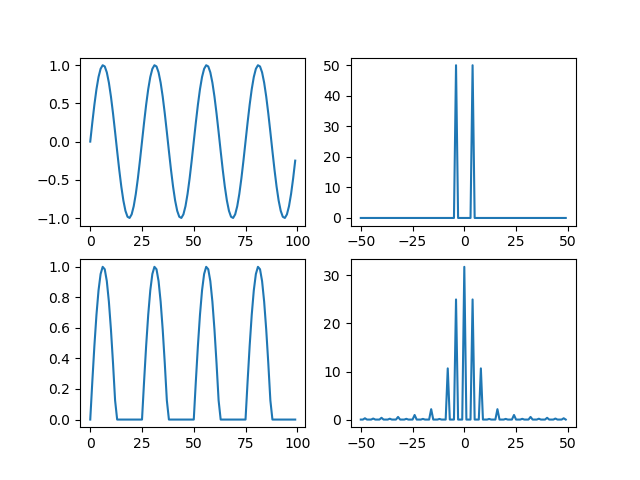}
  \caption{ Left: ReLU operation over a sine wave. Right: equivalent Discrete Fourier Transform. The ReLU operation spreads the sine wave over the spectrum. }
  \label{fig_sine_relu}
\end{figure}

Figure \ref{fig_sine_relu} top left illustrates an input sine wave with frequency equals to $\mu_1 = 4$ and its DFT in top right.
The two bottom graphs illustrate the ReLU function application in the sine wave and its DFT.

The ReLU function reduces the first harmonic amplitude $F(\mu_1=4)$, adds a DC component $F(\mu_0=0)$ and adds the original sine even harmonics, $F(\mu_2=8)$, $F(\mu_4=16)$, $F(\mu_6=24)$ and so on.

\begin{figure}[t]
  \centering
  \includegraphics[width=\columnwidth]{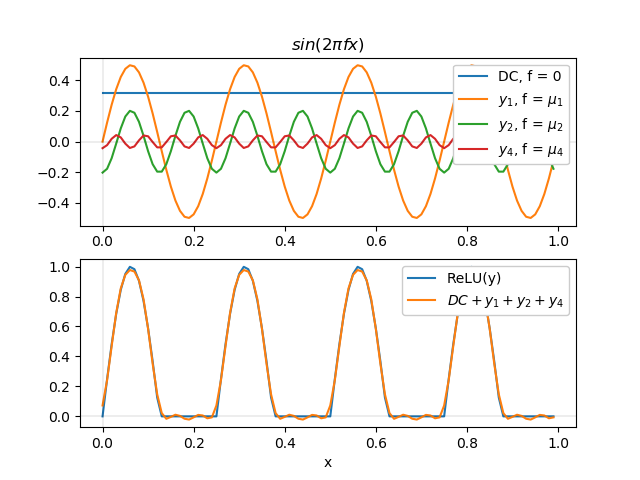}
  \caption{ Top: DC value and the three first sine even harmonics. Bottom: harmonics summation. The number of harmonics determine the approximation quality. }
  \label{fig_series_relu}
\end{figure}

The ReLU operation spreads the input signal in even harmonics over the spectrum.
To illustrate this observation, we can add these frequency components in its original domain.
Figure \ref{fig_series_relu} represents the first 4 frequency components $F(\mu_0=0)$, $F(\mu_1=4)$, $F(\mu_2=8)$, $F(\mu_4=16)$ and their summation (bottom).
Adding more harmonics improve the approximation.

\begin{figure}[t]
  \centering
  \includegraphics[width=\columnwidth]{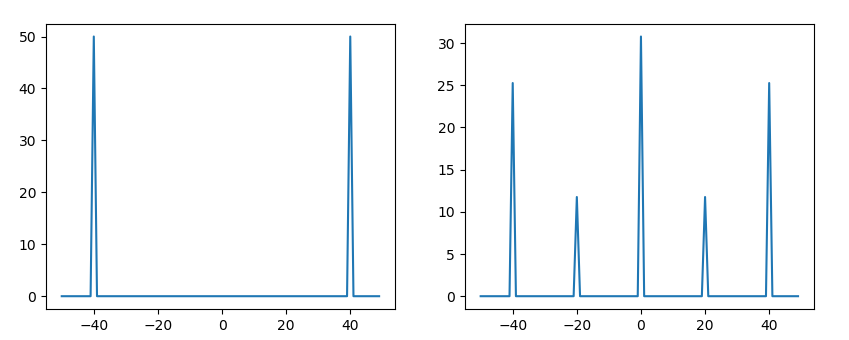}
  \caption{ Sine wave DFT and ReLU-ed sine wave DFT. Sine frequency higher than $F_s/4$. } 
  \label{fig_hf_sine_relu}
\end{figure}

This result is valid when the input sine frequency is lower than one fourth the sampling frequency $\mu_1 < F_s/4$. 
When the sine frequency is higher than $Fs/4$ the DFT displays a different result.
Figure \ref{fig_hf_sine_relu} illustrates the DFT of a ReLU-ed sine wave with frequency $\mu_1 = 40$.
In this example the sine second harmonic, $\mu_2$, is 80 but since this value is higher than the Nyquist frequency, $\mu_{max} = 50$, its DFT presents aliasing.

Since we are generating the input data, increasing the sampling frequency to 200, $F_s=200$, would solve the problem.
But then the same problem would occur for a ReLU-ed sine with frequency higher than 100.

Generally speaking, to compute a ReLU-ed sine DFT the sine frequency must be lower than $F_s/4$.
In intervals higher than $F_s/4$ the ReLU operation will add lower frequency components to the signal.






This example was meant to provide a general view of the ReLU layer operation in the frequency domain.
This evaluation provided the intuition of our 2SReLU layer presented in Section \ref{superposition_srelu}.




\subsection{Layers Disposition}

The \textbf{convolution/batch normalization/ReLU} (CBR) is the main building block for convolutional neural networks.
Analyzing these layers in the frequency domain is possible to notice some reasons for this configuration effectiveness.

The convolutional layer is the most important layer and it holds most of the network learnable parameters (weights).
Its equivalent in the frequency domain is the straightforward point-wise multiplication.
The other layers function is to guide and ease the convolutional layer weights adjustment and consequently the network optimization.

\begin{figure}
  \centering
  \includegraphics[width=1.1\columnwidth]{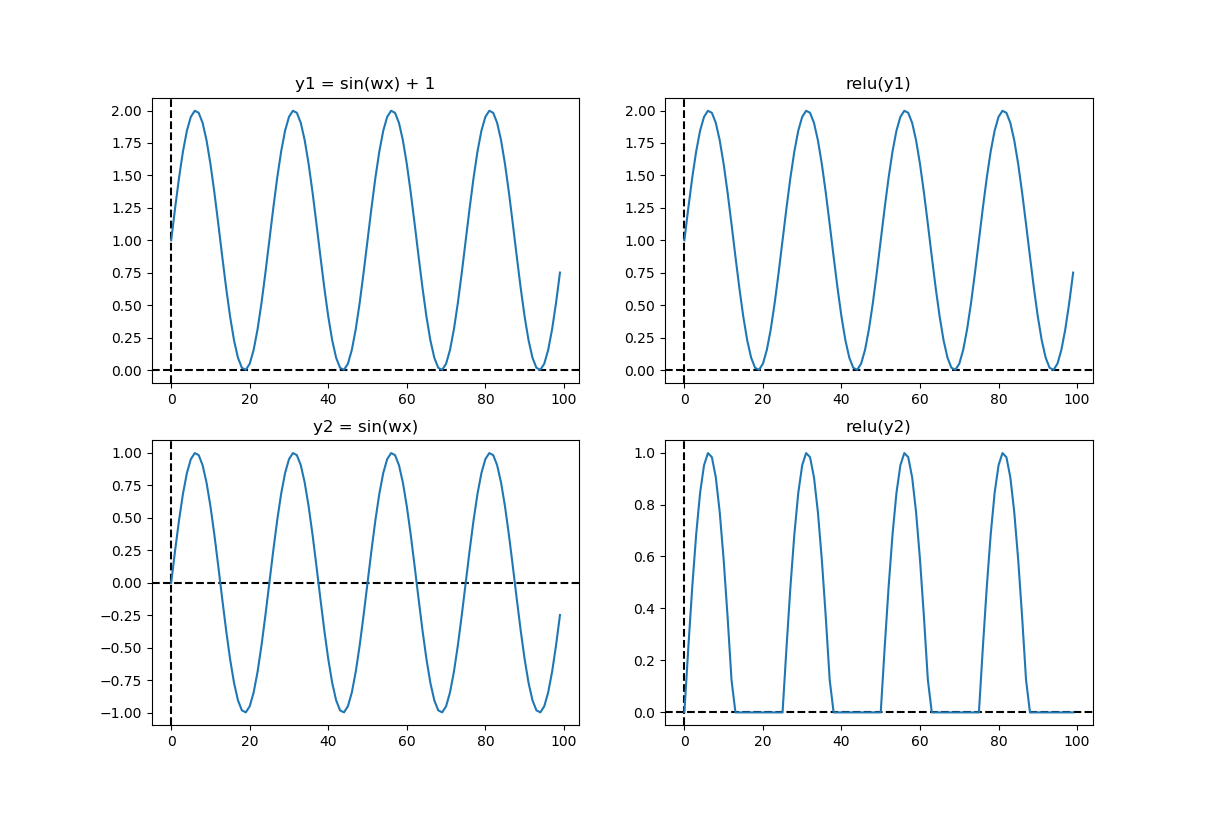}
  \caption{The ReLU layer operation only removes non-positive values. Top: non-negative signals are not modified by the ReLU layer. Bottom: zero centered input signals potentializes the ReLU operation. }
  \label{bn_and_relu}
\end{figure}

The batch normalization has two main purposes.
The batch normalization removes the feature maps mean value and normalizes them to have unit variance.
Removing the feature maps mean value improves the ReLU operation that will be able to add ``non-linearities'', illustrated in Figure \ref{bn_and_relu}.
While bounding the feature maps to unit variance provides numerical stability to the optimization process once the feature maps and consequently the weights will present smaller numbers.

In deep CNNs, the feature maps resolution is decreased and this reduction directly implies in high frequency components removal.
This is noticeable in the spectral pooling layer, which the sole purpose is to remove high frequency components.
Removing high frequencies might guide the network to learn more abstract representations, but at the same time these components carry information that might be important depending on the network task.
For instance, image classification does not suffer from high frequency components removal, but tasks that require a precise pixel localization like object detection or image segmentation may be considerably affected by the resolution reduction.

\section{Second Harmonics SReLU}
\label{superposition_srelu}


From the frequency analysis, in Section \ref{sec_freq_analysis}, it is possible to observe the ReLU operation spreads the signal frequencies through the spectrum.
Low frequency components are split in even harmonics while high frequency components interfere in lower frequency components.

We propose a spectral ReLU that adds low frequency components with their second harmonics, according to Equation \ref{eq_srelu_definition}.
This equation has two hyperparameters $\alpha$ and $\beta$ utilized to adjust each frequency contribution to the final result.

\begin{equation}
  F(\mu_1) \leftarrow \alpha F(\mu_1) + \beta F(\mu_2)
  \label{eq_srelu_definition}
\end{equation}

This operation is valid for the low-frequencies interval defined in Equation \ref{eq_low_freq}.
The second harmonics from frequencies above this interval will not be computed.

\begin{equation}
  0 < \mu_{low} \leq floor(\mu_{max}/2)
  \label{eq_low_freq}
\end{equation}

Therefore, high frequencies components from the interval defined in Equation \ref{eq_high_freq} will not be updated by the spectral ReLU.

\begin{equation}
  floor(\mu_{max}/2) < \mu_{high} \leq \mu_{max}
  \label{eq_high_freq}
\end{equation}

\begin{figure}
  \centering
  \begin{subfigure}[b]{0.4\columnwidth}
    \centering
    \includegraphics[width=0.8\columnwidth]{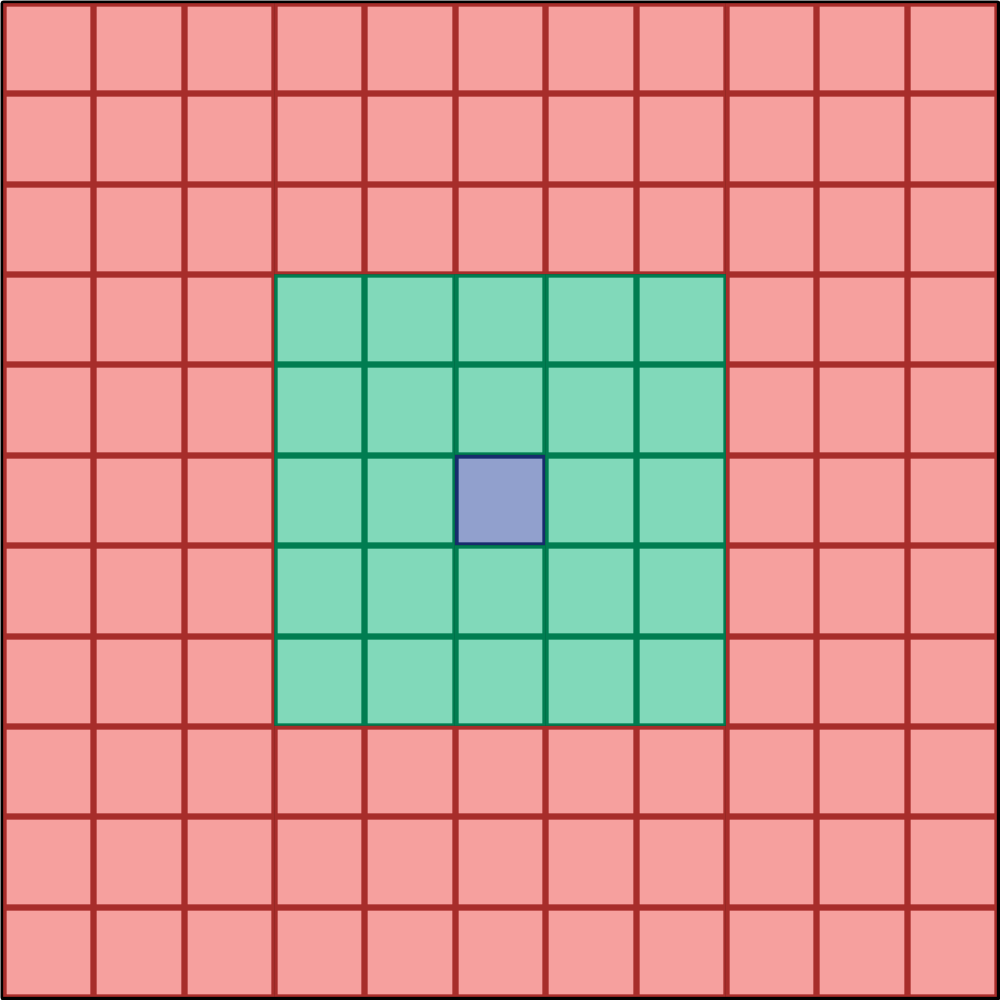}
  \end{subfigure}
  \begin{subfigure}[b]{0.4\columnwidth}
    \centering
    \includegraphics[width=0.8\columnwidth]{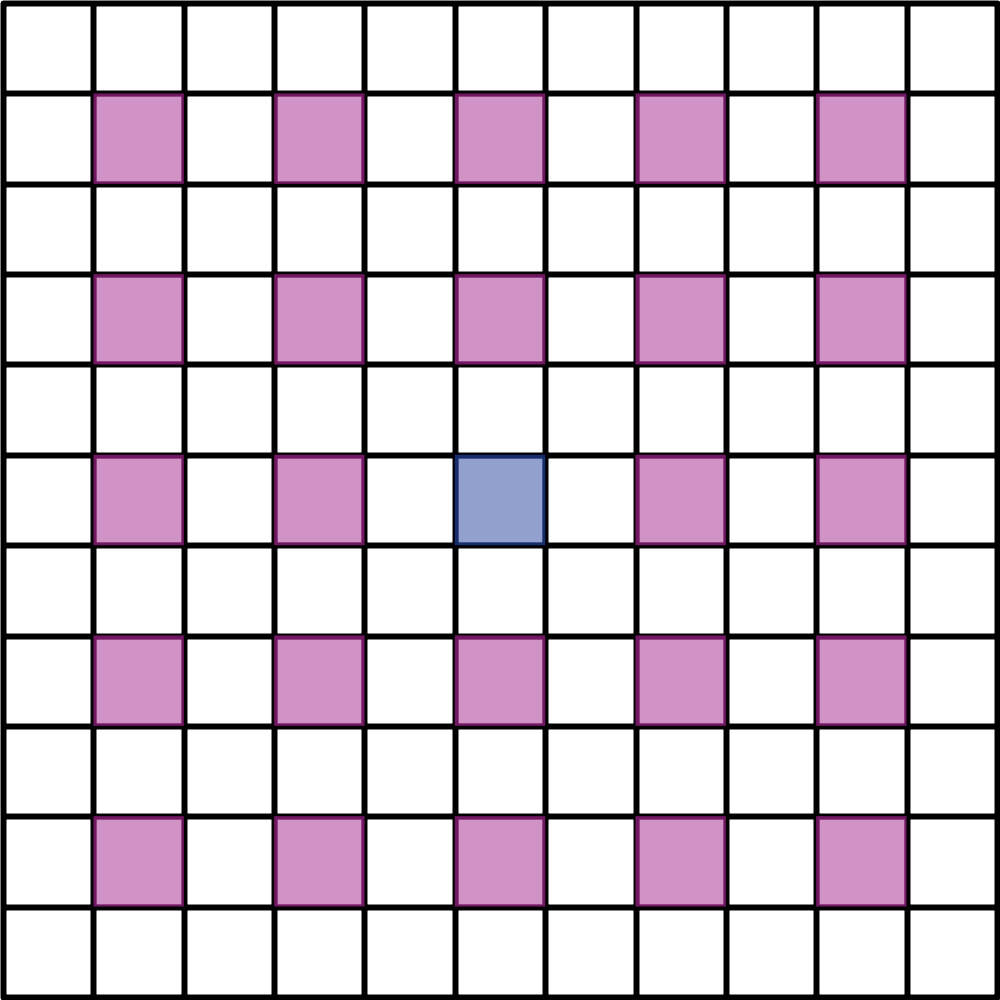}
  \end{subfigure}
  \caption{ 11x11 matrix with centered zero frequency component (blue). Left matrix: 2SReLU computation interval (green). Right matrix: second harmonics (purple). }
  \label{fig_srelu_intervals}
\end{figure}

These two intervals are illustrated in Figure \ref{fig_srelu_intervals}.
The Figure \ref{fig_srelu_intervals} also depicts the second harmonics (purple) that will be used to update the low frequency components.

The 2SReLU Equation \ref{eq_srelu_definition} is the network forward pass equation.
Since this layer has no learnable parameters (weights) there is no update equation and the gradients must be directly propagated.
The 2SReLU backpropagation equation is simply computed by Equation \ref{eq_srelu_backprop}.

\begin{align}
  \label{eq_srelu_backprop}
  \text{grad}_{out}(F(\mu_2)) &\leftarrow \\ \nonumber
      \beta . \text{grad}_{in}&( F(\mu_1) ) + \left< \text{grad}_{in}(F(\mu_2)) \right> \\
  \text{grad}_{out}(F(\mu_1)) &\leftarrow \alpha . \text{grad}_{in}( F(\mu_1) ) \nonumber
\end{align}

$\text{grad}_{in}( F(\mu_1) )$ is the first harmonic input gradient and $\text{grad}_{in}(F(\mu_2))$ is the second harmonic input gradient.
It is important to note that $\text{grad}_{in}(F(\mu_2))$ does not exist it the gradients are coming from the spectral pooling layer.

The proposed spectral ReLU addresses the spectral pooling high frequency components removal applying an operation that preserves this information.

\section{Experimental Results}

To evaluate the proposed spectral ReLU layer a simple convolutional neural network was implemented in the spatial and frequency domain.
To build a frequency domain network, not only the spectral ReLU was manually implemented but the convolution and pooling layers were also manually implemented in the PyTorch framework \cite{paszke2017automatic}.

The classification network was evaluated in two datasets: the MNIST dataset \cite{lecun2010mnist} and the AT\&T face recognition dataset \cite{samaria1994parameterisation}.
These datasets are standard image classification datasets and since they present small images, they would not require a high computational power to validate basic concepts.
Also, we would be able to compare with the SReLU proposed by Ayat \etal \cite{ayat2019spectral}.


\subsection{Convolution in Frequency Domain}

The convolution operation in the frequency domain is equivalent to the pointwise product of the input signal and the convolutional filter.
Both, input signal and convolutional filter (convolutional weights or kernel) are real numbers that must be transformed to the frequency domain, but the Fourier transform of real values are complex numbers.
These complex numbers are usually represented in a rectangular notation, according to Equation \ref{eq_rectangular_notation}, where $i$ is the imaginary number.

\begin{align}
  \label{eq_rectangular_notation}
  x &= x_1 + i x_2 \\
  w &= w_1 + i w_2 \nonumber
\end{align}

The product of two complex numbers in rectangular notation is given by \ref{eq_rectangular_notation_prod}.
Note the result requires 6 operations: 4 multiplications and 2 additions.

\begin{align}
  \label{eq_rectangular_notation_prod}
  x \cdot w &= (x_1 + i x_2) \cdot (w_1 + i w_2) \\
  x \cdot w &= (x_1 w_1 - x_2 w_2) + i (x_1 w_2 + x_2 w_1) \nonumber
\end{align}




It is also possible to represent complex numbers in polar notation, Equation \ref{eq_polar_notation}, where $A$ is the magnitude and $\phi$ is the phase.

\begin{align}
  \label{eq_polar_notation}
  x &= A_x \left[cos(\phi_x) + i \cdot sin(\phi_x) \right] = A_x \angle{ \phi_x } \\
  w &= A_w \left[cos(\phi_w) + i \cdot sin(\phi_w) \right] = A_w \angle{ \phi_w } \nonumber
\end{align}

The product of two complex numbers in polar notation is simpler than in rectangular notation.
Basically, the multiplication is given by two operations: the product of the magnitudes and the addition of the phases, Equation \ref{eq_polar_notation_product}.
Therefore, the polar notation presents a lower computational cost than the rectangular notation.

\begin{align}
  \label{eq_polar_notation_product}
  w \cdot x &= A_w \angle{ \phi_w } \cdot A_x \angle{ \phi_x } \\
  w \cdot x &= A_w A_x \angle{ \phi_w + \phi_x} \nonumber
\end{align}

In this work the complex variables are represented in polar notation.
However, during the experiments it was observed that multiplying both magnitudes and phases had the same effect than multiplying magnitudes and adding phases.
Since the same operation (multiplication) was applied in all feature maps, there was no overhead to separate channels and the computation time was lower.
This approximation might be caused by the normalization layers that keep feature maps and weights small.

\subsection{Normalization in Frequency Domain}

As previously explained, usual normalization layers provide numerical stability reducing the feature maps signal amplitude and they also increase the ReLU layer efficiency.

In the frequency domain network we have employed the batch normalization layer (BN), similarly to Ayat \etal \cite{ayat2019spectral}.
Independently of the domain, normalizing the feature maps by the standard deviation regularizes and provides numerical stability to the network.
This low input signal amplitude has also simplified the complex numbers pointwise multiplication.

The mean value removal is equivalent to the DC component removal in the frequency domain.
A layer to directly remove the DC component was also devised but the experiments have demonstrated this operation does not improve the results.
Since the mean value in the frequency domain is a single frequency component, the network can easily minimize its influence during optimization.

\subsection{Accuracy Correspondence}

This experiment evaluates if the frequency domain network accuracy is equivalent to its correspondent spatial domain network.
Theoretically, if both networks in frequency or spatial domain are equivalent, they should present the same accuracy.

First, a baseline network was implemented in the spatial domain then each layer was replaced by its frequency domain representation.
Both network architectures are represented in Table \ref{tab_eval_networks} where the convolutional layer equivalent in the frequency domain was denominated sparse layer.

\begin{table*}
\centering
\captionsetup{hypcap=false}
\captionof{table}{MNIST dataset evaluation with spatial domain network architecture and its equivalent frequency domain network. C stands for convolutional and FC for fully connected (dense).}
\label{tab_eval_networks}
\resizebox{2\columnwidth}{!}{
\begin{tabular}{ c | c c c c c }
  \hline
  \textbf{Space}& & & \textbf{Frequency} & & \\
  \hline
  C/BN/ReLU     &  Sparse/BN/2SReLU  & Sparse/BN/2SReLU  & Sparse/BN/2SReLU  & Sparse/BN/2SReLU  & Sparse/BN/2SReLU \\
  C/BN/ReLU     &  -                & -                & -                & -                & -\\
  Max Pooling   &  Spectral pooling & Spectral pooling & Spectral pooling & Spectral pooling & Spectral pooling \\
  \hline
  -             &                   &                  & concat\_layer    &                  & \\
  \hline
  C/BN/ReLU     &                   &                  & Sparse/BN/2SReLU  &                  & \\
  C/BN/ReLU     &                   &                  & -                &                  & \\
  Max Pooling   &                   &                  & Spectral pooling &                  & \\
  \hline
  FC1           & & & FC1     & & \\
  FC2           & & & -       & & \\
  FC3           & & & -       & & \\
  Softmax       & & & Softmax & & \\
  \hline
\end{tabular}
}
\end{table*}

Deep learning models present several convolutional layers with small kernels.
Differently from these models, the frequency domain network architecture presented a smaller number of layers than its equivalent in the spatial domain.
The proposed experiments suggest that spatial networks are deep and narrow while frequency domain networks are shallow and wide.

The baseline spatial domain network had two convolution blocks (2 x C / BN / ReLU) before each max pooling.
Each factored convolution in the spatial domain was replaced by a single sparse block in the frequency domain.
Since the sparse layer encompasses all frequency components, factored convolutions are pointless.
Factored convolutions in the frequency domain were not able to improve the network performance.

To simulate the receptive field in the frequency domain network, a pyramidal filter approach was employed in the input image spectrum.
The input image DFT was computed in the whole image and in each quadrant of the image.
In each quadrant, the extracted region of interest was zero-padded to restore the original image size and then it was transformed to the frequency domain with a magnitude and phase representation.
This result might suggest the frequency domain network struggles to notice spatial differences (attention) unless we directly point them out.

After the first sparse layers the feature maps were concatenated and them passed to another sparse layer.
At the end, three fully connected layers were utilized in the spatial domain network but only one fully connected layer was employed in the equivalent frequency domain network.
Both networks utilize the softmax layer to generate the discrete probability density output.
Since the network output is a statistical representation there is no frequency domain equivalent for the fully connected and softmax layers.
This approach is similar to the one utilized by Trabelsi \etal \cite{trabelsi2018deep}.

\begin{figure}
  \centering
  \includegraphics[width=\columnwidth]{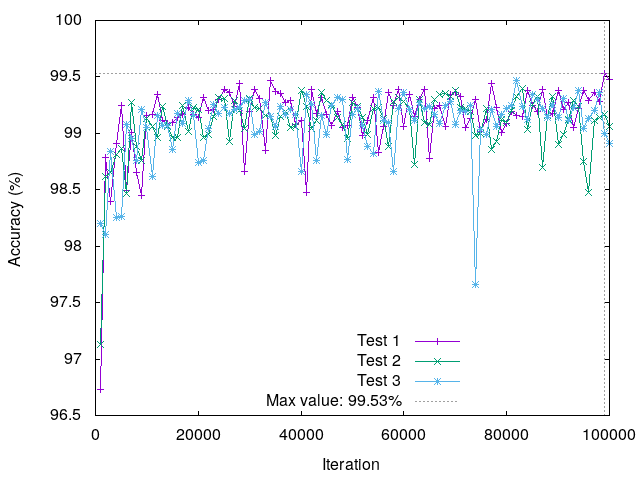}
  \caption{ Convolutional neural network accuracy on MNIST test set for 100k iterations. }
  \label{fig_cnn_network_acc}
\end{figure}

\begin{figure}
  \centering
  \includegraphics[width=\columnwidth]{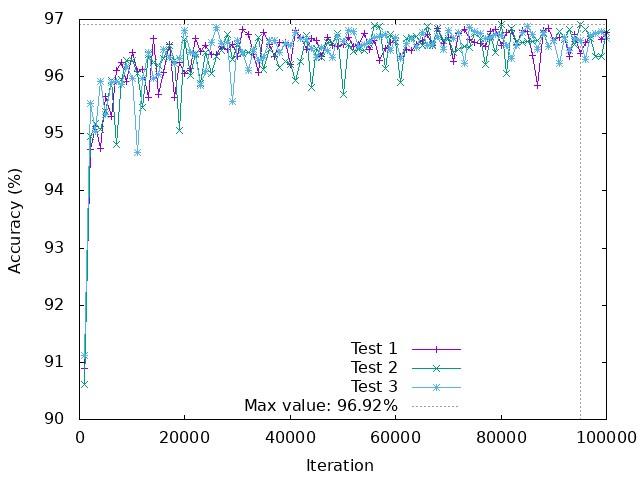}
  \caption{ Spectral neural network accuracy on MNIST test set for 100k iterations. }
  \label{fig_snn_network_acc}
\end{figure}

The experimental results are presented in Figure \ref{fig_cnn_network_acc}, Figure \ref{fig_snn_network_acc} and Table \ref{tab_eval_mnist}.
Figure \ref{fig_cnn_network_acc} and Figure \ref{fig_snn_network_acc} present the accuracy for each network implementation (spatial/frequency).
The Table \ref{tab_eval_mnist} compare the best results achieved for each network in three test runs.

As illustrated in Table \ref{tab_eval_mnist}, our frequency domain network accuracy is close to the spatial domain accuracy, with $2.61\%$ difference.
Ayat \etal \cite{ayat2019spectral} presents a lower accuracy difference, $0.34\%$, but our solution completely removes convolutional layers from the network, while they approximate their spectral ReLU with successive convolutions.
Given the difference between our space and frequency networks, other strategies to preserve high frequency components might improve the results.

\begin{table}
\centering
\captionsetup{hypcap=false}
\captionof{table}{Spatial domain and frequency domain networks accuracy at MNIST dataset.}
\label{tab_eval_mnist}
\resizebox{\columnwidth}{!}
{
\begin{tabular}{ c | c c c }
  \hline
  \textbf{Network type} & \textbf{Ours} & \textbf{Ayat \etal \cite{ayat2019spectral}} & \textbf{Difference} \\
  \hline
  Space domain       & 99.53\%   & 99.69\%    & -0.13\% \\
  Frequency domain   & 96.92\%   & 99.35\%    & -2.43\% \\
  \hline
\end{tabular}
}
\end{table}

The convolutional layer has a high computational cost but a low memory requirement.
In contrast, the sparse layer (or pointwise multiplication) has a lower computational cost but requires a large amount of memory.
For instance, the first convolutional layer had $144$ parameters (weights) while the first sparse layer had $3136$ parameters, an increase factor of $21.78\times$.

To evaluate the 2SReLU contribution the frequency domain network was evaluated without them.
In this case, high frequency components do not influence the network results.
The classification error obtained was 4.42\% and when compared to 3.08\%, the spectral ReLU layers could reduce the error in 1.34\%, a 30\% reduction in the classification error.


The frequency domain network was also evaluated in the AT\&T face recognition dataset \cite{samaria1994parameterisation}.
The evaluated space and frequency domain network architectures are illustrated in Table \ref{tab_eval_network_att}

\begin{table}
\centering
\captionsetup{hypcap=false}
\captionof{table}{AT\&T dataset evaluation with spatial domain network architecture utilized as the baseline and its equivalent frequency domain network. C stands for convolutional and FC for fully connected (dense).}
\label{tab_eval_network_att}
\begin{tabular}{ c c }
  \hline
  \textbf{Space}& \textbf{Frequency} \\
  \hline
  C/BN/ReLU     & Sparse/BN/2SReLU \\
  C/BN/ReLU     & -                \\
  Max Pooling   & Spectral pooling \\
  \hline
  C/BN/ReLU     & Sparse/BN/2SReLU \\
  C/BN/ReLU     & -                \\
  Max Pooling   & Spectral pooling \\
  \hline
  FC1           & FC1 \\
  FC2           & -   \\
  Softmax       & Softmax \\
  \hline
\end{tabular}
\end{table}

The space domain network is the same one utilized in the MNIST dataset.
In this case the pyramidal approach did not improved the results with the frequency domain network.
The faces are stretched along the images and they do not present evident differences in specific positions.
Therefore, extracting regions of interest would not provide meaningful information.

The numerical results are illustrated in Table \ref{tab_eval_att}.

\begin{table}
\centering
\captionsetup{hypcap=false}
\captionof{table}{Space and frequency domain networks accuracy at AT\&T dataset.}
\label{tab_eval_att}
\resizebox{\columnwidth}{!}
{
\begin{tabular}{ c | c c c }
  \hline
  \textbf{Network type} & \textbf{Ours} & \textbf{Ayat \etal \cite{ayat2019spectral}} & \textbf{Difference} \\
  \hline
  Space domain          & 100.00\%      & 98.75\%   & +1.25\%  \\
  Frequency domain      & 97.50\%       & 97.50\%   &  0.00\%  \\
  \hline
\end{tabular}
}
\end{table}

In this dataset our results are on par with Ayat \etal \cite{ayat2019spectral}.
A possible reason for the higher results may be the dataset larger images, the AT\&T dataset present images with higher resolution than the MNIST dataset.
With larger images our spectral ReLU is more effective and retain more high frequency components, given the same network architecture.

\section{Conclusion}

This paper presented the 2SReLU, a novel frequency domain non-linear activation function.
The layer was proposed considering the ReLU layer transformations in the frequency components.
It preserves high frequency components through the network and it is a direct solution to the spectral pooling high frequency components removal.
2SReLU was implemented and tested in a convolution-free frequency domain network presenting competitive results with a lower computational cost than its equivalent spatial network.

As future work, other strategies to superpose frequency components could be devised.
For instance, a mean value of the frequencies around the second harmonics could be utilized.
The layer forward hyperparameters could be replaced by learnable parameters.
Also, an inverse 2SReLU layer that adds the first harmonics to the second harmonics could be formulated.
The inverse 2SReLU layer could be employed before upsample layers (inverse of pooling layers) in frequency domain encoder-decoder networks utilized for image segmentation.

\section*{Acknowledgment}

This research was funded by Sao Paulo Research Foundation (FAPESP), project \#2015/26293-0.
This study was partially financed by Coordenacao de Aperfeicoamento de Pessoal de Nivel Superior (CAPES) - Finance Code 001.


\bibliographystyle{ai_bibtex/IEEEtran}
\bibliography{ai_bibtex/references}

\begin{thebibliography}{10}
\providecommand{\url}[1]{#1}
\csname url@samestyle\endcsname
\providecommand{\newblock}{\relax}
\providecommand{\bibinfo}[2]{#2}
\providecommand{\BIBentrySTDinterwordspacing}{\spaceskip=0pt\relax}
\providecommand{\BIBentryALTinterwordstretchfactor}{4}
\providecommand{\BIBentryALTinterwordspacing}{\spaceskip=\fontdimen2\font plus
\BIBentryALTinterwordstretchfactor\fontdimen3\font minus
  \fontdimen4\font\relax}
\providecommand{\BIBforeignlanguage}[2]{{%
\expandafter\ifx\csname l@#1\endcsname\relax
\typeout{** WARNING: IEEEtran.bst: No hyphenation pattern has been}%
\typeout{** loaded for the language `#1'. Using the pattern for}%
\typeout{** the default language instead.}%
\else
\language=\csname l@#1\endcsname
\fi
#2}}
\providecommand{\BIBdecl}{\relax}
\BIBdecl

\bibitem{krizhevsky2012imagenet}
A.~Krizhevsky, I.~Sutskever, and G.~E. Hinton, ``Imagenet classification with
  deep convolutional neural networks,'' in \emph{Advances in neural information
  processing systems}, 2012, pp. 1097--1105.

\bibitem{ribeiro2016should}
M.~T. Ribeiro, S.~Singh, and C.~Guestrin, ``Why should i trust you?: Explaining
  the predictions of any classifier,'' in \emph{Proceedings of the 22nd ACM
  SIGKDD international conference on knowledge discovery and data
  mining}.\hskip 1em plus 0.5em minus 0.4em\relax ACM, 2016, pp. 1135--1144.

\bibitem{lundberg2017unified}
S.~M. Lundberg and S.-I. Lee, ``A unified approach to interpreting model
  predictions,'' in \emph{Advances in Neural Information Processing Systems},
  2017, pp. 4765--4774.

\bibitem{rudin2019stop}
C.~Rudin, ``Stop explaining black box machine learning models for high stakes
  decisions and use interpretable models instead,'' \emph{Nature Machine
  Intelligence}, vol.~1, no.~5, pp. 206--215, 2019.

\bibitem{rippel2015spectral}
O.~Rippel, J.~Snoek, and R.~P. Adams, ``Spectral representations for
  convolutional neural networks,'' in \emph{Advances in neural information
  processing systems (NIPS)}, 2015, pp. 2449--2457.

\bibitem{wang2016combining}
Z.~Wang, Q.~Lan, D.~Huang, and M.~Wen, ``Combining fft and spectral-pooling for
  efficient convolution neural network model,'' in \emph{2nd International
  Conference on Artificial Intelligence and Industrial Engineering (AIIE2016)},
  2016.

\bibitem{pratt2017fcnn}
H.~Pratt, B.~Williams, F.~Coenen, and Y.~Zheng, ``Fcnn: Fourier convolutional
  neural networks,'' in \emph{Machine Learning and Knowledge Discovery in
  Databases, ECML PKDD}, vol. 10534.\hskip 1em plus 0.5em minus 0.4em\relax
  Springer, 2017.

\bibitem{ayat2019spectral}
S.~O. Ayat, M.~Khalil-Hani, A.~A.-H. Ab~Rahman, and H.~Abdellatef,
  ``Spectral-based convolutional neural network without multiple
  spatial-frequency domain switchings,'' \emph{Neurocomputing}, vol. 364, pp.
  152--167, 2019.

\bibitem{ioffe2015bn}
S.~Ioffe and C.~Szegedy, ``Batch normalization: Accelerating deep network
  training by reducing internal covariate shift,'' in \emph{Proceedings of the
  32nd International Conference on Machine Learning (ICML-15)}, D.~Blei and
  F.~Bach, Eds.\hskip 1em plus 0.5em minus 0.4em\relax JMLR Workshop and
  Conference Proceedings, 2015, pp. 448--456, preprint at
  \url{http://jmlr.org/proceedings/papers/v37/ioffe15.pdf}.

\bibitem{wu2018group}
Y.~Wu and K.~He, ``Group normalization,'' in \emph{Proceedings of the European
  Conference on Computer Vision (ECCV)}, 2018, pp. 3--19.

\bibitem{trabelsi2018deep}
C.~Trabelsi, O.~Bilaniuk, Y.~Zhang, D.~Serdyuk, S.~Subramanian, J.~F. Santos,
  S.~Mehri, N.~Rostamzadeh, Y.~Bengio, and C.~J. Pal, ``Deep complex
  networks,'' in \emph{International Conference on Learning Representations
  (ICLR)}, 2018.

\bibitem{nair2010rectified}
V.~Nair and G.~E. Hinton, ``Rectified linear units improve restricted boltzmann
  machines,'' in \emph{Proceedings of the 27th international conference on
  machine learning (ICML-10)}, 2010, pp. 807--814.

\bibitem{arjovsky2016unitary}
M.~Arjovsky, A.~Shah, and Y.~Bengio, ``Unitary evolution recurrent neural
  networks,'' in \emph{International Conference on Machine Learning}, 2016, pp.
  1120--1128.

\bibitem{guberman2016complex}
N.~Guberman and A.~Shashua, ``On complex valued convolutional neural
  networks,'' \emph{arXiv preprint arXiv:1602.09046}, 2016.

\bibitem{he2016deep}
K.~He, X.~Zhang, S.~Ren, and J.~Sun, ``Deep residual learning for image
  recognition,'' in \emph{Proceedings of the IEEE Conference on Computer Vision
  and Pattern Recognition}, 2016, pp. 770--778.

\bibitem{smith2018discrete}
J.~S. Smith and B.~M. Wilamowski, ``Discrete cosine transform spectral pooling
  layers for convolutional neural networks,'' in \emph{International Conference
  on Artificial Intelligence and Soft Computing}.\hskip 1em plus 0.5em minus
  0.4em\relax Springer, 2018, pp. 235--246.

\bibitem{ryu2018dft}
J.~Ryu, M.-H. Yang, and J.~Lim, ``Dft-based transformation invariant pooling
  layer for visual classification,'' in \emph{Proceedings of the European
  Conference on Computer Vision (ECCV)}, 2018, pp. 84--99.

\bibitem{xu2019dct}
Y.~Xu and H.~Nakayama, ``Dct based information-preserving pooling for deep
  neural networks,'' in \emph{2019 IEEE International Conference on Image
  Processing (ICIP)}.\hskip 1em plus 0.5em minus 0.4em\relax IEEE, 2019, pp.
  894--898.

\bibitem{zhang2018hartley}
H.~Zhang and J.~Ma, ``Hartley spectral pooling for deep learning,'' \emph{arXiv
  preprint arXiv:1810.04028}, 2018.

\bibitem{simonyan2014very}
K.~Simonyan and A.~Zisserman, ``Very deep convolutional networks for
  large-scale image recognition (vgg),'' \emph{arXiv preprint arXiv:1409.1556},
  2014, preprint at \url{https://arxiv.org/abs/1409.1556}.

\bibitem{paszke2017automatic}
A.~Paszke, S.~Gross, S.~Chintala, G.~Chanan, E.~Yang, Z.~DeVito, Z.~Lin,
  A.~Desmaison, L.~Antiga, and A.~Lerer, ``Automatic differentiation in
  pytorch,'' \emph{-}, 2017, available: at \url{https://pytorch.org/}.

\bibitem{lecun2010mnist}
Y.~LeCun, C.~Cortes, and C.~Burges, ``Mnist handwritten digit database,''
  \emph{AT\&T Labs [Online]. Available:
  \url{http://yann.lecun.com/exdb/mnist}}, vol.~2, p.~18, 2010.

\bibitem{samaria1994parameterisation}
F.~S. Samaria and A.~C. Harter, ``Parameterisation of a stochastic model for
  human face identification,'' in \emph{Proceedings of 1994 IEEE Workshop on
  Applications of Computer Vision}.\hskip 1em plus 0.5em minus 0.4em\relax
  IEEE, 1994, pp. 138--142.

\end{thebibliography}

\end{document}